\documentclass[lettersize,journal]{IEEEtran}
\usepackage{amsmath,amsfonts}
\usepackage{algorithmic}
\usepackage{algorithm}
\usepackage{array}
\usepackage[caption=false,font=normalsize,labelfont=sf,textfont=sf]{subfig}
\usepackage{textcomp}
\usepackage{stfloats}
\usepackage{url}
\usepackage{verbatim}
\usepackage{graphicx}
\usepackage[style=ieee, citestyle=ieee-comp, url=false, isbn=false]{biblatex}
\usepackage{xcolor}
\usepackage{booktabs}
\usepackage[margin=10pt]{caption}
\def\decsai{Department of Computer Science and Artificial Intelligence of the University of Granada (DECSAI), Spain}
\def\decsaiShort{DECSAI, Spain}

\def\panacea{Panacea Cooperative Research S. Coop., Ponferrada, Spain}

\def\cespuIHB{Associate Laboratory i4HB - Institute for Health and Bioeconomy, University Institute of Health Sciences (IUCS), CESPU, 4585-116 Gandra, Portugal}

\def\cespuUNIPRO{UNIPRO-Oral Pathology and Rehabilitation Research Unit, IUCS-CESPU, 4585-116 Gandra, Portugal}

\def\cespuUMIB{UMIB-Multidisciplinary Biomedical Research Unit, Abel Salazar Institute of Biomedical Sciences (ICBAS), University of Porto, 4050-313 Porto, Portugal}

\def\cespuUCIBIO{UCIBIO-Research Unit on Applied Molecular Biosciences, Forensic Science Research Laboratory, IUCS-CESPU (1H-TOXRUN), 4585-116 Gandra, Portugal}

\def\momiasBolzano{Institute for Mummy Studies, Eurac Research, Viale Druso 1, 39100, Bolzano, Italy}
\def\udc{Department of Computer Science and Information Technologies, University of A Coru\~na, Spain}
\def\dasci{Andalusian Research Institute in Data Science and Computational Intelligence}

\begin{document}

\title{Automatic dental superimposition of 3D intraorals and 2D photographs for human identification}

\author{Antonio D. Villegas-Yeguas, Xavier Abreu-Freire, Guillermo R-Garc\'ia, Andrea Valsecchi, Teresa Pinho, Daniel P\'erez-Mongiovi, Oscar Ib\'a\~nez, Oscar Cord\'on, \IEEEmembership{Fellow, IEEE}
\thanks{
A. D. Villegas-Yeguas is with \decsai{} and with \panacea{} (e-mail: advy99@correo.ugr.es).
}
\thanks{
Xavier Abreu-Freire and Daniel P\'erez-Mongiovi are with \cespuIHB{} and \cespuUCIBIO{} (e-mail: ruireivax2001@gmail.com; daniel.mongiovi@iucs.cespu.pt).
}
\thanks{
G. R-Garc\'ia is with \panacea{} and with \momiasBolzano{} (e-mail: guiramgar96@gmail.com).
}
\thanks{
Andrea Valsecchi is with \panacea{} (e-mail: andrea.valsecchi@panacea-coop.com).
}
\thanks{
Teresa Pinho is with \cespuUNIPRO{} and \cespuUMIB{} (e-mail: teresa.pinho@iucs.cespu.pt).
}
\thanks{
O. Ib\'a\~nez is with \udc{} and \panacea{} (e-mail: oscar.ibanez@udc.es).
}
\thanks{
O. Cord\'on is with \decsaiShort{} and \dasci{} (e-mail: ocordon@decsai.ugr.es)
}

}

\maketitle

\begin{abstract}
	Dental comparison is considered a primary identification method, at the level of fingerprints and DNA profiling. One crucial but time-consuming step of this method is the morphological comparison. One of the main challenges to apply this method is the lack of \textit{ante-mortem} medical records, specially on scenarios such as migrant death at the border and/or in countries where there is no universal healthcare. The availability of photos on social media where teeth are visible has led many odontologists to consider morphological comparison using them. However, state-of-the-art proposals have significant limitations, including the lack of proper modeling of perspective distortion and the absence of objective approaches that quantify morphological differences.

	Our proposal involves a 3D (\textit{post-mortem} scan) - 2D (\textit{ante-mortem} photos) approach. Using computer vision and optimization techniques, we replicate the \textit{ante-mortem} image with the 3D model to perform the morphological comparison. Two automatic approaches have been developed: i) using paired landmarks and ii) using a segmentation of the teeth region to estimate camera parameters. Both are capable of obtaining very promising results over $20,164$ cross comparisons from $142$ samples, obtaining mean ranking values of $1.6$ and $1.5$, respectively. These results clearly outperform filtering capabilities of automatic dental chart comparison approaches, while providing an automatic, objective and quantitative score of the morphological correspondence, easily to interpret and analyze by visualizing superimposed images.

\end{abstract}


\section{Introduction} \label{introduction}

\IEEEPARstart{D}{isaster} Victim Identification (DVI) scenarios, where a large number of victims need to be identified, have become more common in recent decades. The main causes of these scenarios are military actions, natural disasters, and migratory crisis, situations where it is difficult to find \textit{ante-mortem} data to perform the identification. This is the case of gold standard techniques such as DNA or fingerprint identification. Despite their high reliability, these techniques cannot be applied when the soft tissue does not survive for friction ridge analysis or DNA sequencing, or there is no known sample to compare with.   

The dentition contains the hardest and most resilient tissues in the human body, being resistant to decomposition, high temperatures and other extreme environmental conditions. In addition to these factors, its unique structure and characteristics make the dentition one of the most useful and valuable structures for the task of human identification. For all these reasons, forensic odontology is considered a primary method of identification by Interpol \cite{forrest_forensic_2019, cerritelli_interpol_2023-1}, and have proven to be highly useful in DVI scenarios \cite{perrier_swiss_2006}.

Identification using forensic odontology is performed by comparing a set of \textit{ante-mortem} (AM) records of a missing person with a set of \textit{post-mortem} (PM) records of a cadaver. There are two widely used approaches to perform this comparison: i) morphological comparison, where dental structures are visually compared, and ii) dental record comparison, where descriptions of dental status are compared based on a coding system that describes the possible states of a tooth and its possible anomalies \cite{forrest_forensic_2019}. Usually morphological comparison methods are preferred over comparison of dental records due to their higher reliability, level of confidence, and individualizing power, since the images used are objective records of the dental status of each subject. 

One of the biggest challenges in forensic odontology is the collection of good quality AM data. This data is obtained from the medical records of hospitals, dental clinics, and dental practices. In certain scenarios, such as migrant DVI, it is very difficult to obtain these records due to legal obstacles, socioeconomic conditions, and lack of family records. 

For these reasons, in recent years attempts have been made to develop techniques that allow identification with a type of AM data that is more common and easier to obtain: photographs where the dentition is visible \cite{de_angelis_dental_2007, bollinger_grinline_2009, santoro_personal_2019, reesu_forensic_2020, naidu_exploring_2022, de_sousa_human_2025}. Although all these proposals show how this technique could be of great help in the field of forensic identification, they have certain problems, such as not taking into account all the parameters of the camera or using manual tools that can distort the images when performing the superimposition. In this paper we will approach the problem as a 3D-2D image registration (IR) problem \cite{markelj_review_2012}, proposing two different methods, one based on landmarks and the other based on regions. To validate the proposals we consider 142 pairs of 3D intraoral scans (IOS) of the teeth and facial photographs where the dentition is visible, which we will divide into three partitions based on the visibility of the teeth. We use the error of the registration in pixels to sortlist each comparison, obtaining a ranking from more to least plausible superimposition. With that ranking we compute statistics to measure the performance of the proposals. Additionally, in this work we make the first proposal for the use of likelihood ratio (LR) in the field of dental comparison. The LR \cite{vergeer_specific-source_2023} is a well established framework \cite{van_lierop_overview_2024}, recommended by the European Network of Forensic Science Institutes (ENFSI) \cite{champod_enfsi_2016} as it express subjective probabilities. The use of LR also allows us to better put into context the reliability of the proposed methods.

\section{Background: Dental superimposition in forensic odontology and 3D-2D image registration for superimposition} \label{background}
\noindent

\subsection{Dental superimposition in forensic odontology}

Many different approaches have been proposed for dental superimposition using photographs, testing its potential applicability using both real and simulated cases. One of the first proposals is that of De Angelis et al. \cite{de_angelis_dental_2007}. In this work, the authors propose a preliminary protocol for evaluating dental superimposition. This protocol is based on aligning an AM photograph, in which the teeth are perpendicular to the camera, and a simulated PM photograph of the 3D model of the teeth. To do so, the lowest visible point of both canine teeth and the interdental point between the central incisor are used. Using these three points, both AM and PM photographs are aligned and scaled, and an index of correspondence is computed using the contours of the teeth.

In 2009, Bollinger et al. proposed a methodology to superimpose AM and PM dental images using Adobe Photoshop™ \cite{bollinger_grinline_2009}. The method was validated with PM photographs of ten subjects and corresponding old photographs as AM data.

Another proposal is that of Santoro et al., who used Adobe Photoshop\textsuperscript{TM} and Facecomp\textsuperscript{TM} to perform dental superimposition \cite{santoro_personal_2019}. They used ten photographs of different subjects as AM data, and ten photographs of 3D plaster models as PM data. The proposed method consists of performing a registration of both photographs using five landmarks positioned on the canine and incisor teeth. 

One year later, Reesu et al. explored the feasibility of this method to improve the accuracy rate in identification processes \cite{reesu_forensic_2020}. Three experienced forensic odontologists and three novel forensic odontology MSc. students evaluated $31$ 3D models and $35$ digital photographs using two different approaches, a visual comparison and a 3D-2D superimposition performed manually using the 3D Rhinoceros\textsuperscript{TM} software.

Given the growing interest in this technique, in 2021 Naidu et al. conducted a survey where over 80 forensic odontologists and related professionals were asked about the usefulness of selfies in human identification scenarios \cite{naidu_exploring_2022}. The survey showed that more than $30\%$ of participants already used this type of data, while another $41\%$ planned to use it in the future.

In 2022, Mazur et al. studied the relationship between the distortion of the smile line and the focal length when using photographs to perform human identification \cite{mazur_smile_2022}. With a sample of $28$ persons, they compare one AM photograph with three simulated PM photographs with different focal lengths (18mm, 55mm and 80mm). They proved how the focal length is significant when comparing smile lines in photographs, suggesting the need to properly accounting for perspective distorsion when comparing two dental images.

Recently, in 2025, De Sousa et al. compared two different approaches to perform dental comparison using photographs and 3D models \cite{de_sousa_human_2025}. They used both a comparison of the smile line and a dental superimposition using De Angelis' proposal \cite{de_angelis_dental_2007}, showing how both approaches can be very useful, especially for the exclusion of individuals in identification processes.

These studies highlight the relevance and impact of this technique in forensic scenarios, particularly when AM information is difficult to obtain. However, all of these proposals present problems with regard the data processing or the methodology that can lead to errors. Most rely on manual, subjective comparisons and use small samples, along with several technical issues discussed below. 

The 2D-2D superimposition is only valid in constrained and unrealistic conditions where both AM and PM images have the same pose and perspective distortion. For this reason, the 3D-2D IR has been extensively studied and used in other areas of forensic anthropology, such as skull-face overlay in craniofacial \cite{valsecchi_robust_2018} or comparative radiography \cite{gomez_3d-2d_2018}. Although this technique is more complex in craniofacial superimposition, several recommendations and best practices \cite{damas_handbook_2020} also apply to dental superimposition. These include avoiding modifications such as cropping, and taking into account the perspective of the photograph, among others. We can see how these issues are present in the discussed works, so in our proposal we will focus on addressing them. To do so, our approach is to treat the dental comparison task as a 3D-2D IR problem.

\subsection{The 3D-2D image registration problem}

From the perspective of optics and computer science, this problem has been extensively studied \cite{markelj_review_2012}. The IR problem consists of aligning two images into a common coordinate system, keeping one of them fixed. In the case of 3D-2D IR, the 2D image remains fixed, and the objective is to find the pose and intrinsic parameters of the camera with which the image was taken, a problem also known as the camera calibration problem.

3D pose refers to the position and orientation of an object in a three-dimensional space. Estimating the 3D pose thus involves computing the position and orientation of an object relative to the camera in a 3D scene. There are different approaches for estimating the 3D pose \cite{xu_critical_2024}. One of the best knowns is the Perspective-n-Point ($PnP$) problem, where given $n$ points in 3D space $a_1, \dots, a_n$ and their corresponding points in the projected 2D image $b_1, \dots, b_n$, the pose of a calibrated camera is computed by obtaining a projection $P$ that minimizes the projection error (Equation \eqref{eq:projectionError}):

\begin{equation}
\label{eq:projectionError}
    \frac{1}{n}\sum_{i=1}^{n}{\| P(a_i), b_i \|}
\end{equation}

\noindent where $P(a_i)$ is the projection of the 3D point $a_i$ in 2D and $\| \cdot \|$ is the 2D Euclidean distance. Multiple approaches have been proposed to find the best projection $P$ \cite{lepetit_epnp_2009}. One of the main drawbacks of this approach is that it requires a calibrated camera. For this reason, there are also studies that, in addition to solving the $PnP$ problem, also estimate the intrinsic parameters of the camera \cite{lourakis_model-based_2013}. For the vast majority of cameras, these internal parameters can be simplified to the focal length, and for this reason this problem is known as $PnP+f$, where $f$ refers to the focal length.

Another approach is to use an optimization algorithm to search for the camera parameters that maximizes the overlap between the projection of the region of interest of the 3D model and the object in the 2D image \cite{markelj_review_2012, gomez_3d-2d_2018, hu_segmentation-driven_2019}. This approach has the advantage that it can be used without a set of homologous corresponding points (landmarks), as it only uses the segmentation of the object in the 2D and 3D image. Besides, comparing the silhouette of the anatomical region instead of only a set of landmarks allows the method to have a more informed metric of the superimposition, while also been able to produce an output which is more similar to how the forensic expert performs the morphological comparison. The main drawbacks of this approach is that there is no exact algorithm for finding a solution, we need to explore the solution space to find the best projection, thus the need of a metaheuristic: an algorithm capable of find near-optimal solutions within a reasonable amount of time.

\section{Proposal} \label{proposal}
\noindent

In this paper we propose two approaches to perform identification using photographs and 3D IOS: i) to use landmarks to perform superimpositions by solving the $PnP+f$ problem; ii) to segment the regions of interest in the photographs and 3D models, searching for the camera parameters using a cost-efficient evolutionary algorithm. Both methods will give us a metric measuring the superimposition error. We will use this metric to sort each comparison of the same case and generate a ranking, from most similar to least similar. The comparison of both methods will be done using the statistics obtained by the rankings, explained in details in the following section. We also propose the first LR framework in dental comparison to obtain a measure of how informative the proposed methods are, following the recommendations of ENFSI.

As discussed in Section \ref{background}, in both cases we will try to find the camera parameters to superimpose the 3D model over the 2D dental photograph, using seven parameters in total: translation on the X, Y and Z axis; rotation around the X, Y and Z axis; and focal length.

Since premolars and molars are rarely visible in photographs, only incisors and canines were used in this study.

\subsection{First approach: Superimposition using landmarks} \label{proposalLandmarks}

This proposal is to use the \textit{Posest} algorithm to solve the $PnP+f$ problem \cite{lourakis_model-based_2013}, similar to the solution proposed in \cite{valsecchi_robust_2018}, although in our case we do not have the problem of soft tissue since we are comparing the same bone tissue. 

\begin{figure}[!t]
\centering
	\includegraphics[width=0.4\textwidth]{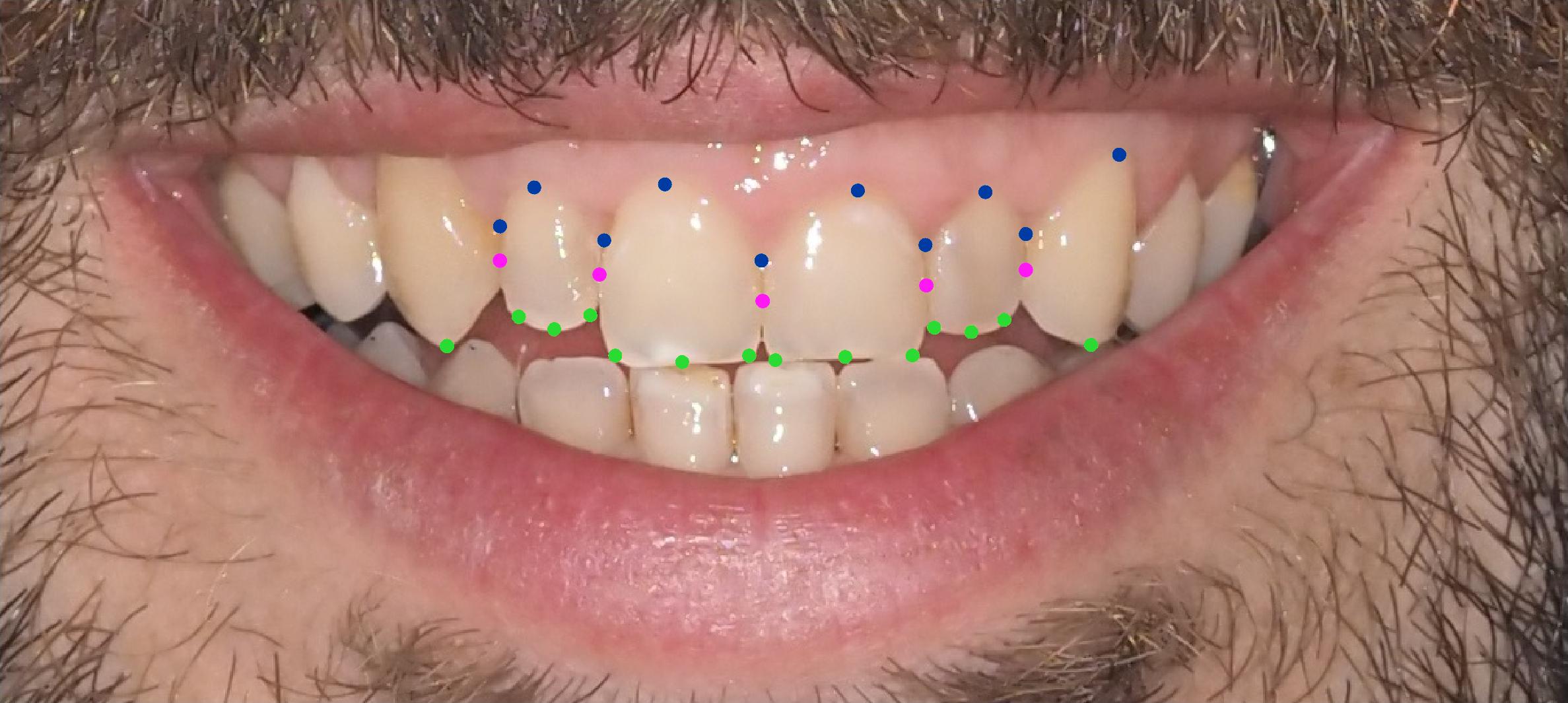}
	\caption{The three landmark sets proposed.}
	\label{proposedLandmarkSets}
\end{figure}

With regard to the landmarks to be used, we propose a whole set including landmarks along the bite, marking the corners of each tooth and its central point; the interdental point in the middle of the teeth; and landmarks at the top of the tooth, next to the gum, as show in Figure \ref{proposedLandmarkSets}.

In these experiments we will also conduct a study of the specific set of landmarks to be used. We will launch the experiments with three different sets of landmarks, selected from the complete set according to their nature:

\begin{itemize}
	\item Set 1: Gingival line, medial line and smile line: 30 landmarks in total. All landmarks in Figure \ref{proposedLandmarkSets}.
	\item Set 2: Medial line and smile line: 19 landmarks in total. Green and purple landmarks in Figure \ref{proposedLandmarkSets}.
	\item Set 3: Smile line: 14 landmarks in total. Green landmarks in Figure \ref{proposedLandmarkSets}.
\end{itemize}

We have differentiated these three sets because this allows us to study how the use of landmarks in the gingival and medial areas affects the results. These landmarks provide more information for superimposition, but they can be altered by pathologies related to gingival problems \cite{macri_periodontal_2024}.

The aim of this study is to find the minimal set of landmarks providing a good performance, i.e. a set of landmarks that are quick to place, with robustness and accuracy, while yielding reliable identification results. Finding a landmark set that meet these requirements is of great importance. 

To measure how good a superimposition is, we will use the the root-mean-square error (RMSE) (Equation \eqref{eq:rootProjectionError}), to order each comparison from most to least similar. Since this error is measured on the projected image, it will be measured in pixels.

\begin{equation}
\label{eq:rootProjectionError}
    \sqrt{\frac{1}{n}\sum_{i=1}^{n}{\| P(a_i), b_i \|}}
\end{equation}

When using the RMSE of the projection to compare superimpositions we encounter a problem regarding the scale of images. If we compare two superimpositions of the same 3D model projected over two different photographs, each photograph can have a different resolution, or the size of the dentition in the photograph can be different. The RMSE is relative to that resolution and scale, so it would be erroneous to directly compare the two values of RMSE. To deal with this problem, when performing the comparisons we will compare each AM photograph against all the PM 3D IOS, to ensure that the size and scale of the dentition is fixed for each case ranking.

\subsection{Second approach: Superimposition using regions}

The region-based approach is based on segmenting the teeth area in both the 2D photograph and the 3D model. The goal is to find the camera parameters used to take the 2D photograph, so that the projection of the 3D segmentation using those camera parameters matches the segmentation of the teeth in the 2D photograph.

This parameter estimation task is an optimization problem with an enormous solution space, making it impossible to find the optimal solution for each superimposition. A common way to find good solutions in this type of scenario is to use metaheuristics: optimization and solution search algorithms capable of exploring the solution space in a way that allows finding a solution that, although without guarantees of optimality, yields good results. These types of algorithms have proven to yield very good results in a wide variety of problems over the years \cite{gendreau_handbook_2019}.

Another detail to consider is the execution time of these algorithms. They are guided by a function that evaluates how good or bad a solution is, and in our case, where we need to render and compare 2D images, these fitness functions are computationally expensive. This leads us to the need to use cost-efficient metaheuristics that are capable of finding good solutions with a low number of evaluations of the function to be optimized.

For all these reasons we propose using a variant of the Mean Variance Mapping Optimization evolutionary algorithm (MVMO-SH) \cite{rueda_hybrid_2013}, to find the camera parameters that will give us the best superimposition. MVMO-SH has proven to be extremely useful for problems where evaluating a solution is a very time-consuming operation \cite{camargo_comparison_2014, gomez_performance_2020} thanks to its rapid convergence mechanisms. 

\begin{figure}[!t]
\centering
	\includegraphics[width=0.4\textwidth]{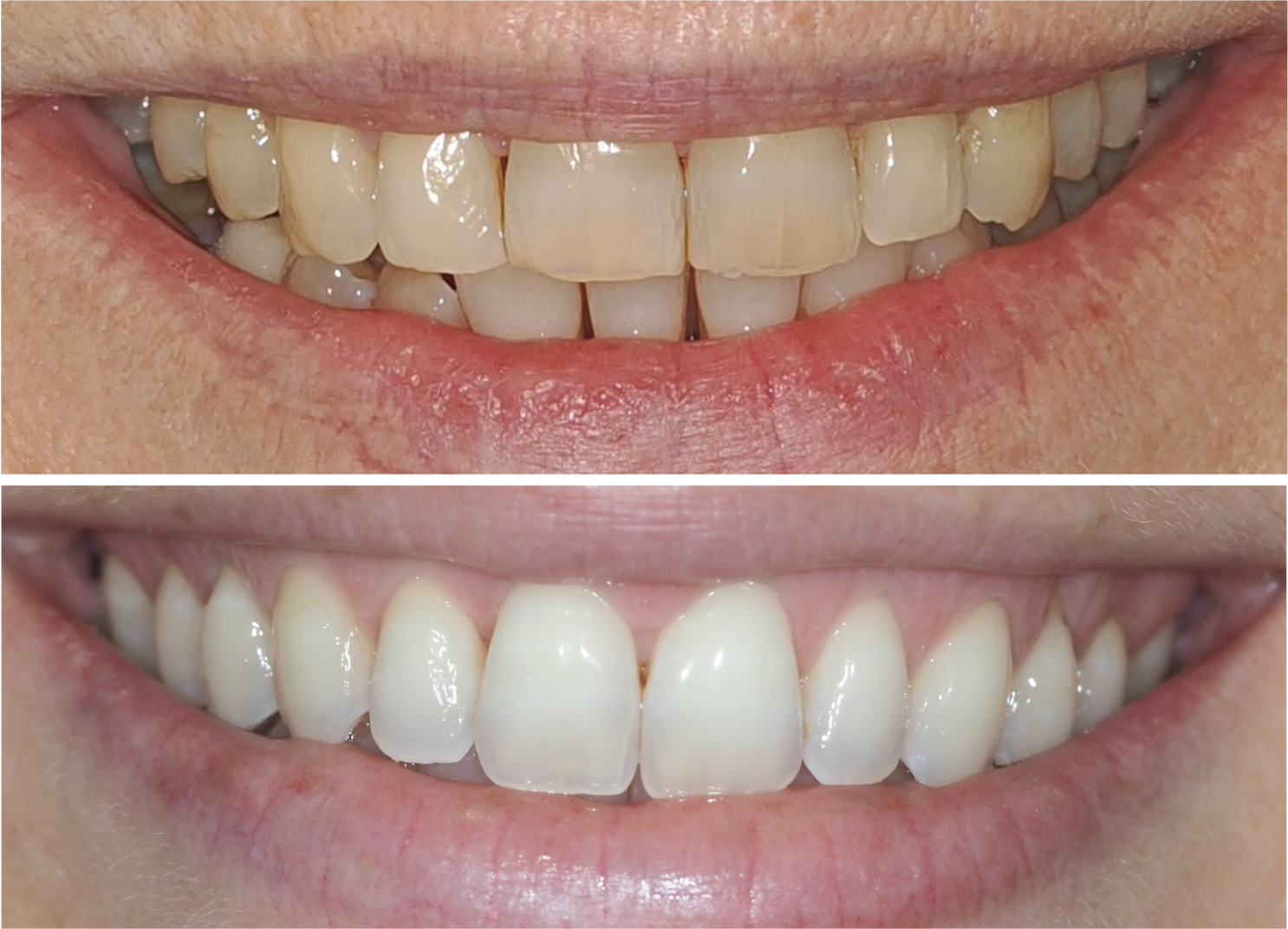}
	\caption{Examples of occlusion of the upper part (top image) and lower part (bottom images) of teeth in the 2D photos due to the lips.}
	\label{exampleOcclusionProblems}
\end{figure}

\begin{figure}[!t]
\centering
	\includegraphics[width=0.4\textwidth]{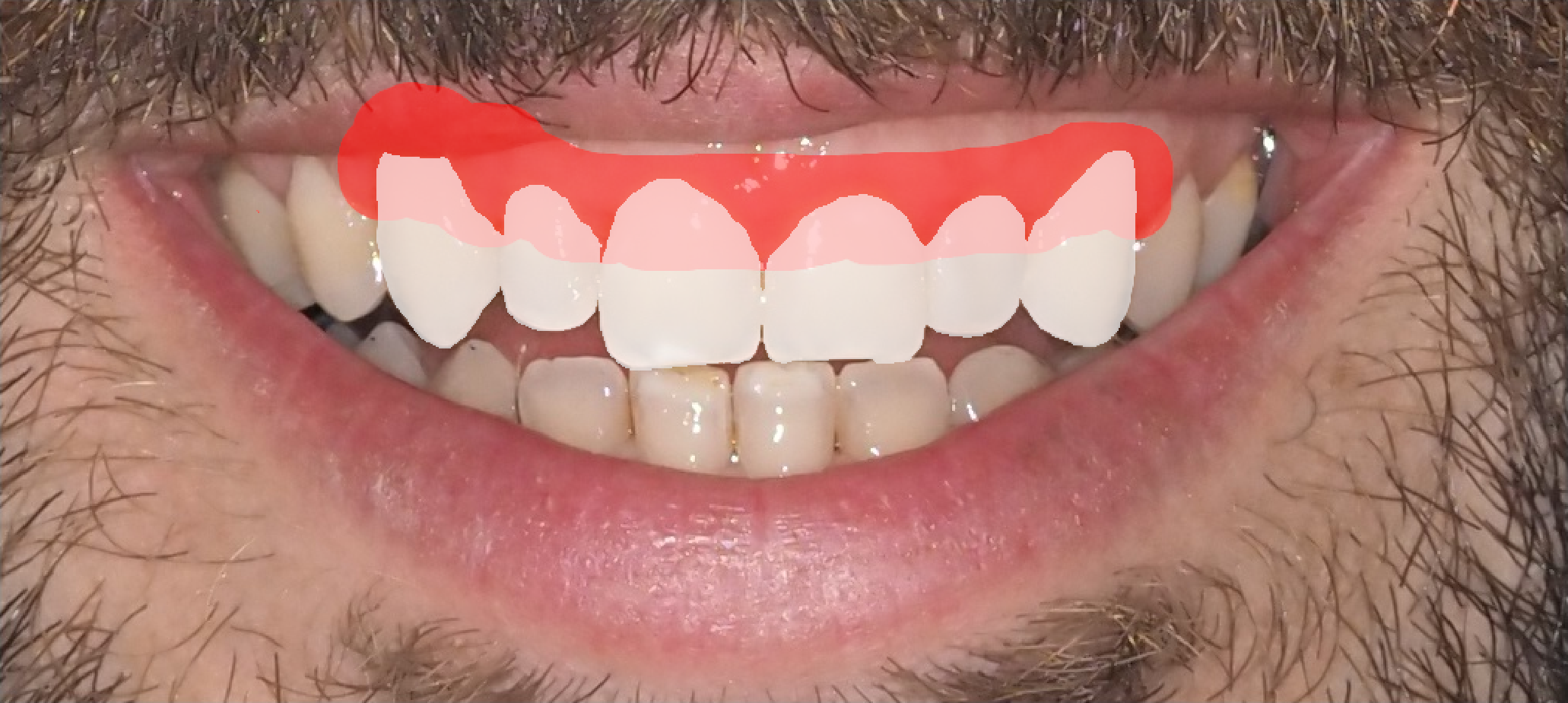}
	\caption{Example of the segmentation for one case. In white the region of interest and in red the occlusion mask.}
	\label{example2Dsegmentation}
\end{figure}

One issue to be considered in this approach is that the teeth region may be in occlusion due to the lips, as shown in Figure \ref{exampleOcclusionProblems}, in addition to possible alterations in the gingival tissue, either due to pathology or loss of soft tissue after death. To take this into account, our fitness function will be the classical DICE metric, but masking areas of possible occlusion or that may be modified between the AM and PM data collection, as done in \cite{gomez_3d-2d_2018, gomez_performance_2020} (see Equation \eqref{eq:MaskedDICE}):

\begin{equation}
\label{eq:MaskedDICE}
    \text{MaskedDICE} = \frac{2 \cdot | (I_a \setminus M) \cap (I_b \setminus M) | }{ | I_a \setminus M | + | I_b \setminus M | }
\end{equation}

\noindent where $I_a$ is the segmentation of the teeth in the photograph, $I_b$ is the projection of the segmentation of the 3D IOS, and $M$ is the occlusion mask. An example of this approach can be seen in Figure \ref{example2Dsegmentation}. The masked DICE values goes from zero to one, where one is a perfect overlap. As the MVMO-SH algorithm minimizes the error, we will invert the error range, so a masked DICE of zero is a perfect overlap.

We will use the masked DICE values to order comparisons and thus create a ranking, from best to worst superimpositions. We still face the same problem of size of the photographs and scale of the dentition. To deal with this, we follow the same approach described at the end of the previous subsection.

\section{Experiments and analysis of results} \label{experiments}
\noindent

\subsection{Experimental design}

For this study, three different sets of 3D IOS and photograph pairs were selected, depending on the visibility of the teeth in the photographs. The first set consists of $50$ cases where the teeth are fully visible, making it easy to see the smile and gingival line. The second set consists of $50$ cases where there is some occlusion in the smile or gingival line, making the teeth not fully visible. Finally, the third set consists of $42$ cases where there is clearly some occlusion problem, with part of the teeth or smile line not visible. This data was selected so that the AM photograph was taken between one and three years prior to the 3D scan. Overall we have 142 pairs of 3D IOS and photographs, obtained from the \textit{Clínica Médico-Dentária de São João da Madeira, Lda}, Portugal, with the approval from the Ethics Committee of the University Institute of Health Sciences (CESPU), reference 19/CE-IUCS/2021. To clarify the reading of this text, we will refer to each dataset as follows:

\begin{itemize}
    \item Dataset A: 50 cases where the teeth are fully visible. 
    \item Dataset B: 50 cases where the teeth have some occlusion. 
    \item Dataset C: 42 cases where part of the teeth are clearly in occlusion.
    \item Complete dataset: The 142 cases in datasets A, B, and C.
\end{itemize}

For the landmark-based approach we will conduct an experiment for each dataset (A, B, C and complete) with each of the landmark sets (Sets 1, 2, and 3, see Section \ref{proposalLandmarks}), 12 experiments in total. Both 2D and 3D landmarks have been placed using the commercial software Skeleton-ID \cite{valsecchi_skeleton-id_2023}, as it allows us to work with both 3D and 2D images, with specific tools for placing landmarks.

\begin{figure}[!t]
\centering
	\includegraphics[width=2.75in]{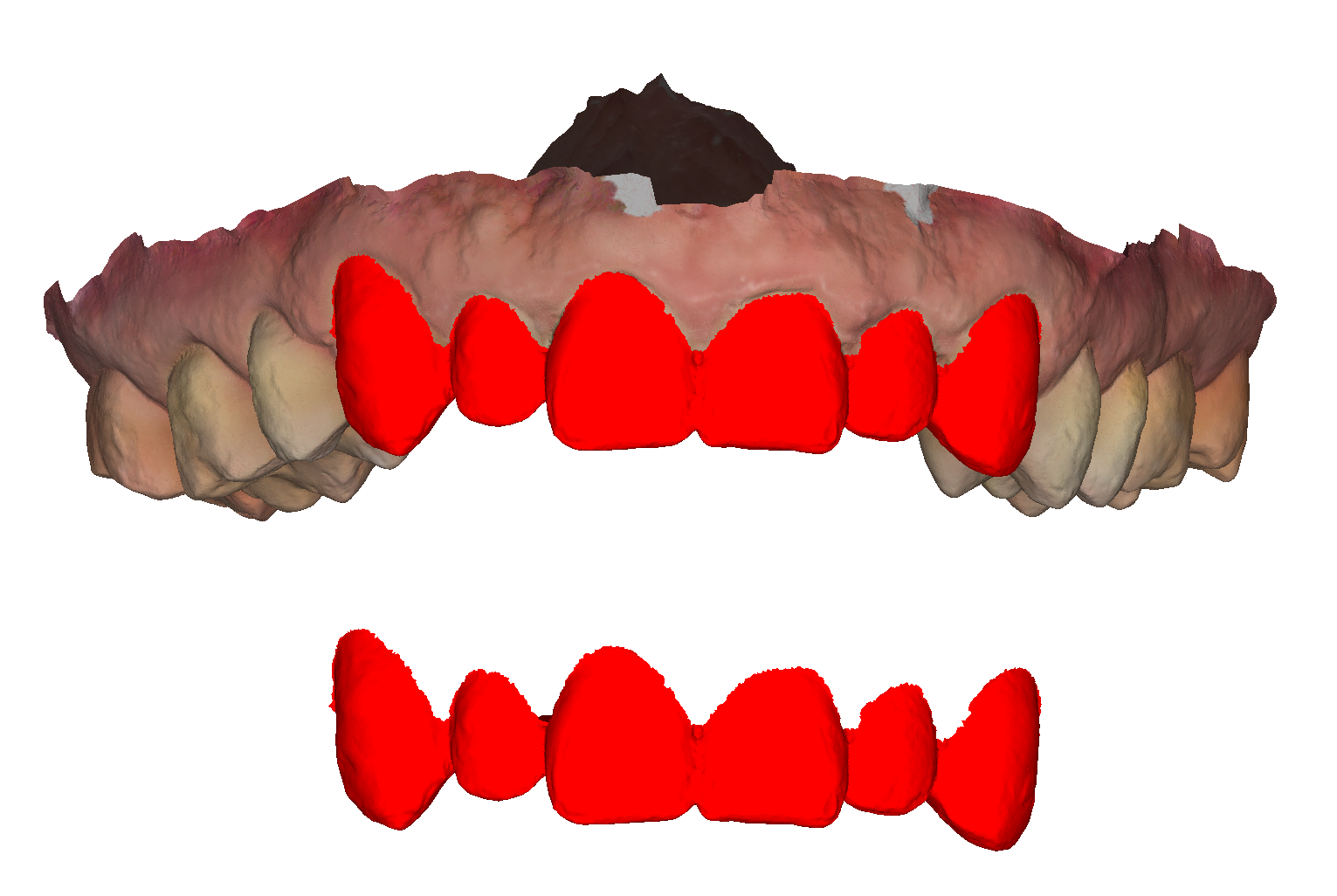}
	\caption{Example of a 3D mesh segmentation. Above the original 3D mesh with the region of interest colored in red and below the final 3D mesh used for the comparison.}
	\label{example3Dsegmentation}
\end{figure}

Regarding the region-based approach, we will conduct four experiments, one for each dataset. In this case, the photographs have been segmented using the free and open-source software GNU Image Manipulation Program. The 3D IOS have been segmented using the open-source software SculptGL, drawing the vertices and faces of interest and making a selection by color, as shown in Figure \ref{example3Dsegmentation}.

In this case, we need to establish a range of values for each parameter to be optimized by MVMO-SH. We have considered the following ranges for each of the seven parameters:

\begin{itemize}
    \item Translation on the X, Y and Z axis: $[-150, 150]$, in millimeters.
    \item Rotation around the X, Y and Z axis: $[-90, 90]$, in degrees.
    \item Focal length: $[10, 200]$, in millimeters.
\end{itemize}

When using an evolutionary algorithm we also need to specify the number of generations. In our case, we have empirically established this number to $600$. The rest of the hyperparameters are those established in the MVMO-SH proposal \cite{rueda_hybrid_2013}, as they have yielded good results. This approach, being a heuristic search method, does not guarantee that we will obtain the optimal solution. For this reason and to avoid problems related to search stagnation, for each comparison the algorithm will be run three times, each one with a different random seed, and the best run will be considered as the final result.

To compare the two approaches we will use the ranking statistics. We also use the LR framework to study how informative are the proposed methods. Both techniques will be introduced in the following two subsections.

\subsection{Ranking statistics}

As said, we will use the metric of each approach to generate a ranking that sorts each comparison from more to less plausible. This ranking is generated for each AM case, obtaining a similarity profile against all PM cases.

With these rankings we can obtain the position within the ranking where the actual AM case matches the PM case, i.e. the correct comparison. This ranking position can take a value between $1$ (the first element of the ranking is the correct comparison) and $N$ (the last element is the correct comparison), where $N$ is the number of PM cases. Ideally we want to find a method that gives us a position of $1$ in every ranking.

In order to study and compare the different experiments, we will use these statistics:

\begin{itemize}
	\item{Average ranking position: On average, number of positions to find the correct comparison.}
	\item{Minimum/maximum ranking position: Of all rankings, the best/worst correct position.}
	\item{Q1/Q2/Q3: Highest value of the correct position within 25\%/50\%/75\% of cases with the lowest correct position.}
	\item{P95/P99: Highest value of the correct position within 95\%/99\% of cases with the lowest correct position.}
\end{itemize}

\subsection{Likelihood ratio} \label{proposal_lr}

In this work we propose the first LR framework in the field of dental comparison. The LR is a statistic that allows us to compare the probabilities of two competing hypotheses, $H_0$ and $H_1$, given some evidence $E$ and conditioned on prior information $I$ \cite{vergeer_specific-source_2023}.

\begin{equation}
\label{eq:LikelihoodRatio}
	\text{LR} = \frac{ p( E | H_0, I) }{p( E | H_1, I) }
\end{equation}

As seen in Equation \eqref{eq:LikelihoodRatio}, LR gives us a value that indicates whether $H_0$ is more likely than $H_1$ or \textit{vice versa}, depending on whether the LR value is greater than $1$ or close to $0$. An LR value close to one indicates that the evidence $E$ equally supports the probability of $H_0$ and $H_1$ conditioned to the prior information $I$.

Multiple metrics have been proposed to evaluate how well an LR system works. The most widely used proposal is the log-likelihood-ratio cost, $C_{llr}$, a metric that tells us the information needed on average to determine the true hypothesis given a set of LR values when we have no \textit{a priori} information. This metric is described in Equation \eqref{eq:CLLR}:

\begin{equation} \label{eq:CLLR}
	\begin{split}
		C_{llr} = \frac{1}{2}(\frac{1}{N_{H_0}}\sum_{i=1}^{N_{H_0}}{log_2(1 + \frac{1}{LR_{H_{0_i}}})} \\
		+ \frac{1}{N_{H_1}} \sum_{j=1}^{N_{H_1}}{log_2(1+LR_{H_{1_j}})})
	\end{split}
\end{equation}

The LR has become a standard statistic in forensic science as it can express the subjectivity and uncertainty associated with certain evidence, and therefore evaluating its strength \cite{van_lierop_overview_2024}. On the one hand, the ENFSI officially recommends using the LR as the standard framework for evaluating and reporting the probative value of forensic evidence, which ensures that forensic conclusions are balanced, logical, robust, and transparent \cite{champod_enfsi_2016}. On the other hand, based on the LR values, we can compare different proposals even when they use different evaluation metrics by calculating the log-likelihood-ratio cost ($C_{llr}$). This applies to comparison of the two approaches proposed in this work, which are evaluated with different metrics. Reporting the reliability of a system based on the $C_{llr}$ goes further, as it allows us to compare different forensic techniques for human identification.

\subsection{Results of the landmark-based superimposition approach}

\begin{table}
\centering
\resizebox{\columnwidth}{!}{%
\begin{tabular}{ccccccccccc}
Landmark Set & Dataset & \# Cases & AVG  & MIN & Q1  & Q2  & Q3  & P95  & P99  & MAX \\ \hline
Set 1        & A       & 50       & \textbf{1.12} & 1   & 1.0 & 1.0 & 1.0 & 2    & 2    & \textbf{2}   \\
Set 2        & A       & 50       & 1.16 & 1   & 1.0 & 1.0 & 2.0 & 2    & 3    & 4   \\
Set 3        & A       & 50       & 1.24 & 1   & 1.0 & 1.0 & 2.0 & 2    & 5.51 & 6   \\ \hline
Set 1        & B       & 50       & \textbf{1.18}& 1   & 1.0 & 1.0 & 1.0 & 2.10 & 4.53 & \textbf{6}   \\
Set 2        & B       & 50       & \textbf{1.18} & 1   & 1.0 & 1.0 & 1.0 & 2.10 & 4.53 & \textbf{6}   \\
Set 3        & B       & 50       & 1.58 & 1   & 1.0 & 1.0 & 1.0 & 5.55 & 9.55 & 12  \\ \hline
Set 1        & C       & 42       & 1.16 & 1   & 1.0 & 1.0 & 1.0 & 2    & 2    & \textbf{2}   \\
Set 2        & C       & 42       & \textbf{1.14} & 1   & 1.0 & 1.0 & 1.0 & 2    & 2.59 & 3   \\
Set 3        & C       & 42       & 1.43 & 1   & 1.0 & 1.0 & 1.0 & 2    & 8.49 & 13  \\ \hline
Set 1        & ALL     & 142      & \textbf{1.60} & 1   & 1.0 & 1.0 & 2   & 5    & 8    & \textbf{9}   \\
Set 2        & ALL     & 142      & \textbf{1.60} & 1   & 1.0 & 1.0 & 2   & 4    & 9    & 14  \\
Set 3        & ALL     & 142      & 2.40 & 1   & 1.0 & 1.0 & 2   & 7    & 24.6 & 58 
\end{tabular}%
}
\caption{Ranking statistics of all the experiments using the landmark-based approach.}
\label{tab:allStatisticsLandmarks}
\end{table}

As can be seen in Table \ref{tab:allStatisticsLandmarks}, the landmark set 1 (the most informed one, as it has 30 landmarks instead of the 19 and 14 landmarks of the sets 2 and 3 respectively) performs better than the rest in both average and maximum ranking, except when used with the dataset C, where the landmark set 2 is slightly better in average ranking, but not in maximum ranking. This difference is more pronounced with dataset A, the dataset with all teeth visible. In datasets B and C, where there are occlusion problems, landmarks close to the gingival line have not been placed in some cases because those regions are not visible, so this behavior was expected. We can also observe that using only the smile line landmarks (landmark set 3) the results are much worse regardless of the dataset used. This is also to be expected, as solving the $PnP+f$ problem using coplanar points is usually more complex and tends to yield worse results \cite{chatterjee_algorithms_2000}.

If we group the results by landmark, the differences per dataset can be identified. The results obtained with dataset A are the best, followed by dataset C and finally dataset B. When using the landmark set 2 the best result is obtained with dataset C, but the worst still corresponds to dataset B. 

\begin{figure}[!t]
\centering
	\includegraphics[width=3.0in]{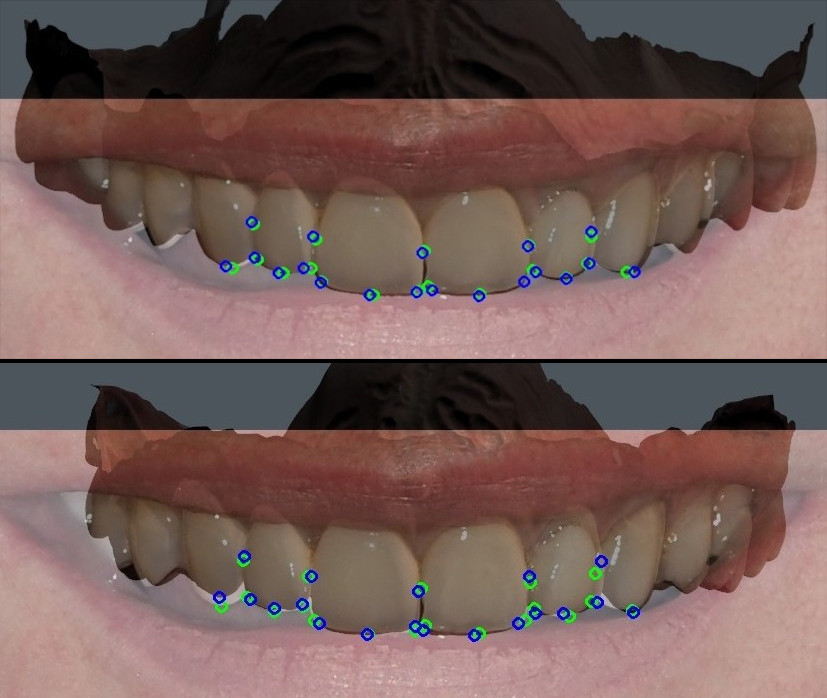}
	\caption{Comparison of the superimposition of the PT0230 photograph with the PT0230 3D IOS (above), and the PT0040 3D IOS (below). The first one with a RMSE of $7.49$ pixels and the second one with a RMSE of $6.87$.}
	\label{exampleSimilarRMSE}
\end{figure}

Looking at the overlaps in detail, positive comparisons that are not in the first position of the ranking are mainly because of occlusion in some landmarks on the canine teeth or incisors, as observed in Figure \ref{exampleSimilarRMSE}. This leads to having multiple landmarks that fit on the same 2D photo with a low RMSE.

Leaving aside the comparison between different experiments, we can see how effective this approach is when it comes to rank the different comparisons. The results show how useful it is for filtering candidates and finding the correct comparison much more quickly. Even in the scenario with the highest number of cases, $142$, the average number of comparisons to check in order to find the correct match is less than two ($1.1\%$ of all the possible cases), and only nine comparisons ($6.34\%$ of the cases) are needed for the worst case.

\subsection{Results of the region-based superimposition approach}

\begin{table}
\centering
\resizebox{\columnwidth}{!}{%
\begin{tabular}{cccccccccc}
Dataset & \# Cases & AVG  & MIN & Q1 & Q2 & Q3 & P95 & P99 & MAX \\ \hline
A       & 50       & \textbf{1}    & 1   & 1  & 1  & 1  & 1   & 1   & \textbf{1}   \\
B       & 50       & 1.24 & 1   & 1  & 1  & 1  & 2   & 6.1 & 7   \\
C       & 42       & \textbf{1}    & 1   & 1  & 1  & 1  & 1   & 1   & \textbf{1}   \\
All     & 142      & 1.5  & 1   & 1  & 1  & 1  & 1   & 15  & 26 
\end{tabular}%
}
\caption{Ranking results of all the experiments using the region-based approach.}
\label{tab:RegionsRanking}
\end{table}

In Table \ref{tab:RegionsRanking} we can see how with this approach we obtain a perfect ranking for datasets A and C, which results in a superb performance. Even so, for dataset B we have some very few cases where the positive superimposition did not have the lowest masked DICE value.

\begin{figure}[!t]
\centering
	\includegraphics[width=3.0in]{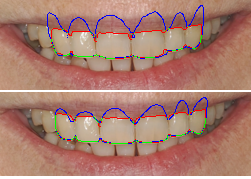}
	\caption{Comparison of the superimposition of the PT0123 photograph with the PT0123 3D IOS (above), and the PT0040 IOS (below). The first one with a masked DICE of $0.051$ and the second one with a masked DICE of $0.031$. In red the silhouette of the 2D teeth, in blue the silhouette of the projected 3D teeth and in green the overlap between the silhouettes.}
	\label{exampleSimilarDentition}
\end{figure}

\begin{figure}[!t]
\centering
	\includegraphics[width=3.0in]{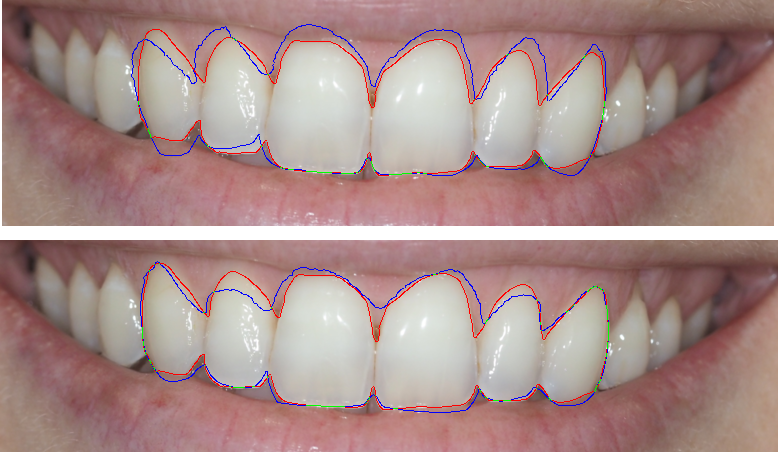}
	\caption{Comparison of the superimposition of the PT0046 photograph with the PT0046 3D IOS (above), and the PT0237 3D IOS (below). The first one with a masked DICE of $0.045$ and the second one with a masked DICE of $0.034$. In red the silhouette of the 2D teeth, in blue the silhouette of the projected 3D teeth and in green the overlap between the silhouettes. }
	\label{exampleLipsOcclusion}
\end{figure}

By reviewing the superimpositions we can identify two scenarios in which the top position in the ranking is not the positive comparison. The first scenario is for cases where the dentition does not have any anomalies or individualizing traits, so several superimpositions obtain a low masked DICE score, e.g. the superimposition shown in Figure \ref{exampleSimilarDentition}. The other scenario involves cases where the teeth are partially covered by the lips, causing the same problem as in the first scenario, as Figure \ref{exampleLipsOcclusion} shows. When processing the AM data, we do not know the silhouette of the dentition as it is occluded by the lower lip, so any 3D IOS can fit in this case, resulting in a good superimposition.

These results also show the great potential of dental superimposition for short listing candidates for identification. By using this approach we get very promising results, with the correct comparison being the first in the ranking in 95\% of the 142 cases.

\subsection{Comparison of approaches: Likelihood ratio and rankings}

As mentioned in Section \ref{introduction}, we can distinguish two main techniques within dental comparison: odontogram comparison and morphological comparison based on images. The aim of this section is to put the results obtained by our proposal into context. However, in the absence of public data, it is impossible to directly compare results, and all we can do is report results from other approaches, based on the metrics they have used taking into account the size and type of the dataset used. 

In the field of morphological comparison, the few existing studies have used samples ranking from $100$ to $207$ cross-comparisons, reporting results based on different metrics, such as a custom index of correspondence, correlation coefficients computed by landmarks, or by the accuracy of the practitioner to find the correct match. Regarding the results of these studies, they report an accuracy between $80\%$ and $93\%$ when the identification is performed by an expert practitioner using the smile line. 

In the field of odontogram comparison, the number of publications is even smaller. In \cite{adams_computerized_2016}, they measure the effectiveness of their proposal on a set of $400$ synthetic samples (only the AM odontogram is available, the PM odontogram was generated based on statistical changes), and, although they do not report results in terms of ranking positions, they are able to find the correct match for the $91\%$ of cases looking only at the first $5\%$ of positions. More recently, in \cite{villegas-yeguas_use_2026}, average ranking results of $1.94$ are reported on a test set of $42$ samples. The metrics and sample size used in this study are very similar to those in our proposal, and serve at least to put into context the results obtained. In this work we improve the filtering capacity of the work using odontograms obtaining a perfect ranking with the datasets of $50$ cases. Thus, our proposals clearly outperform filtering capabilities of published automatic dental chart comparison methods while providing an automatic objective and quantitative score of the morphological correspondence, easy to interpret and analyze by visualizing superimposed images.

However, as we began explaining in this subsection, there are no public datasets that allow for a fair comparison of different past or future proposals. As an alternative, in this paper we have adapted a proposal for calculating the LR \cite{martinez-moreno_evidence_2024} to the specific case of comparing dental morphology using images, as discussed in Section \ref{proposal_lr}. In our specific case, the two competing hypotheses are:

\begin{itemize}
	\item{$H_0$: The subject of the 2D photography and the subject of the 3D IOS are the same.}
	\item{$H_1$: The subject of the 2D photography and the subject of the 3D IOS are different.}
\end{itemize}

\begin{figure}[!t]
\centering
	\includegraphics[width=3.0in]{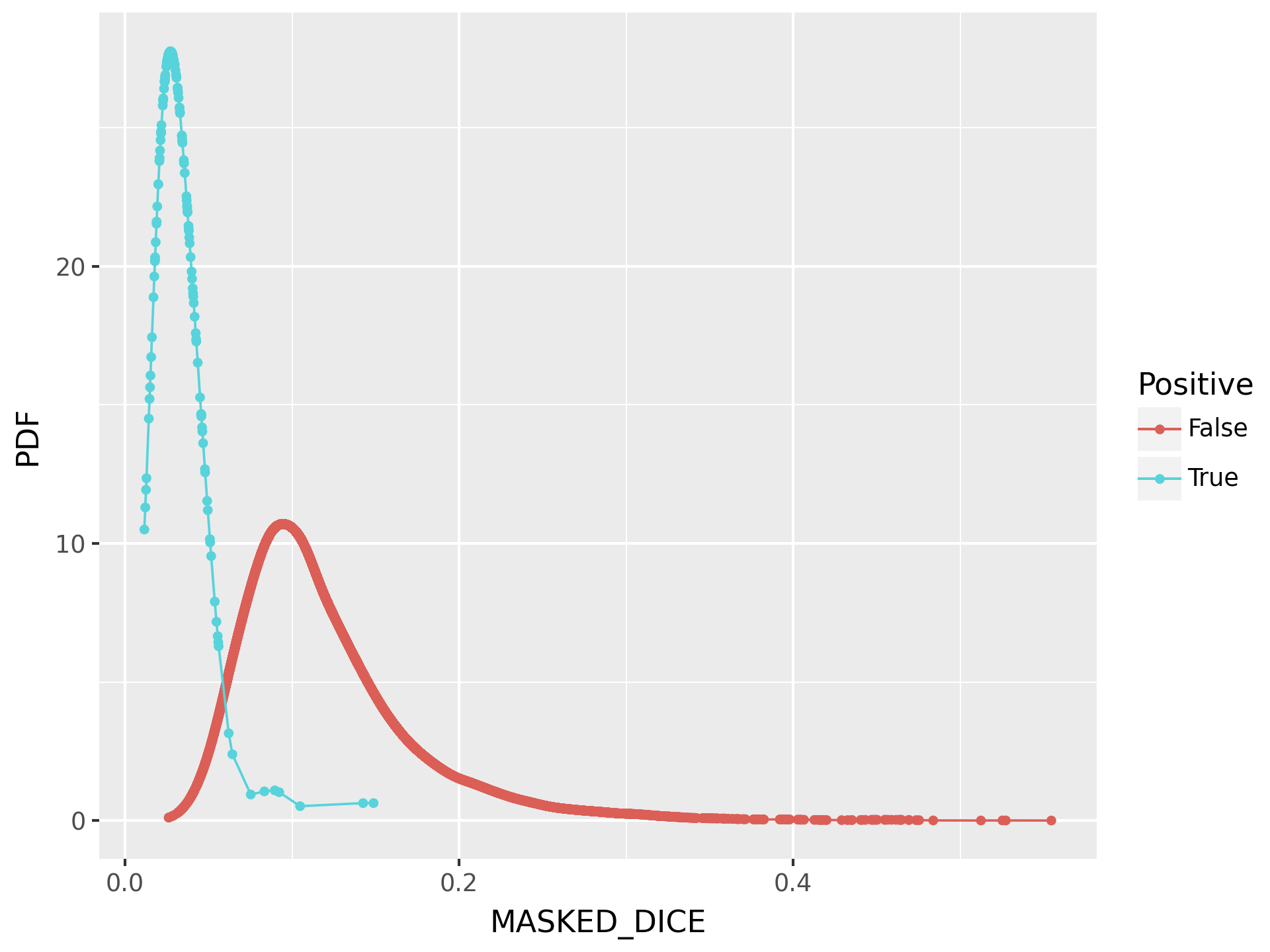}
	\caption{Example of the PDF for the results of the regions approach using the 142 available subjects.}
	\label{PDFExample}
\end{figure}

To compute the LR as in Equation \eqref{eq:LikelihoodRatio}, the evidence $E$ is the score obtained from a comparison, the RMSE for the landmark-based approach and the masked DICE for the region-based approach. To estimate the probability density function (pdf) $p$ we have used a Gaussian kernel density estimation using two sets of scores, obtained by positive comparisons and negative comparisons respectively. Figure \ref{PDFExample} shows an example, visualizing the pdf curves obtained for the region-based approach with all the available data. For each experiment we have computed the corresponding PDFs and obtained the LR values. With those LR values we have also computed the $C_{llr}$, shown in Table \ref{tab:CLLR}, in order to compare the different experiments. 

\begin{table}
\centering
\resizebox{\columnwidth}{!}{%
\begin{tabular}{ccccc}
                                          & Landmark Set 1 & Landmark Set 2 & Landmark Set 3 & Regions         \\ \hline
\multicolumn{1}{c|}{Dataset A (50 cases)} & \textbf{0.122} & 0.219          & 0.231          & 0.166           \\
\multicolumn{1}{c|}{Dataset B (50 cases)}  & 0.266          & 0.266          & 0.315          & \textbf{0.113}  \\
\multicolumn{1}{c|}{Dataset C (42 cases)} & 0.307          & 0.248          & 0.310          & \textbf{0.236}  \\
\multicolumn{1}{c|}{All (142 cases)}      & 0.290          & 0.286          & 0.281          & \textbf{0.2316}
\end{tabular}%
}
\caption{$C_{llr}$ results of all the experiments. The lower, the better.}
\label{tab:CLLR}
\end{table}

\begin{figure}[!t]
	\centering
		\includegraphics[width=3.5in]{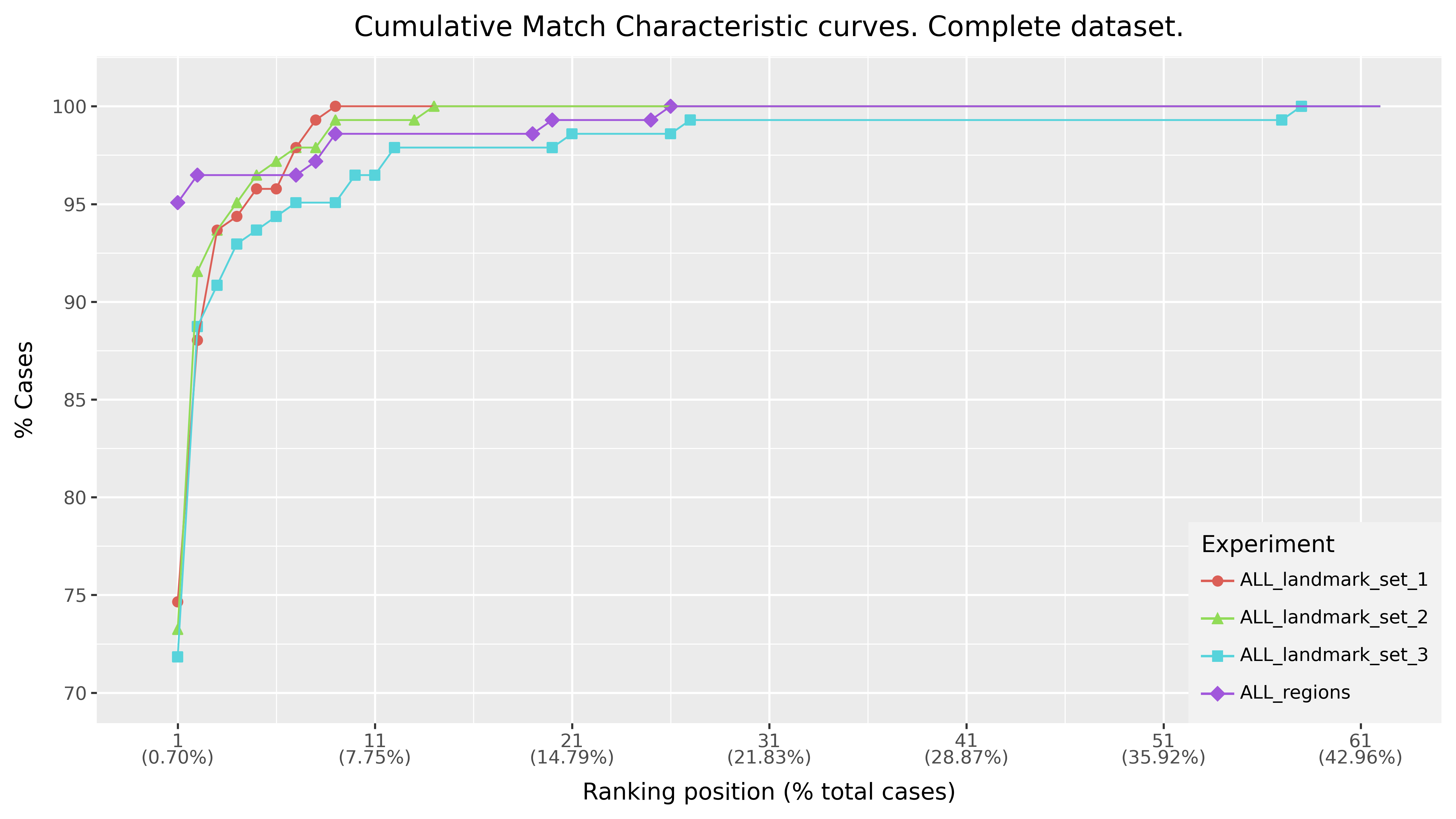}
	\caption{CMC curves for the complete dataset. Note that the Y-axis starts at $70\%$ of cases for a clearer visualization.}
	\label{cmcCurvesALL}
\end{figure}

In Figure \ref{cmcCurvesALL} we also show the cumulative match characteristic (CMC) curves \cite{li_handbook_2011} of the rankings to visualize the behavior of the ranking in the complete dataset.

The results in Table \ref{tab:CLLR} show how, according to the $C_{llr}$, the region-based approach performs better than the landmark-based approach, except for Dataset A. If we compare which model is better according to the $C_{llr}$ and according to the rankings, we see that there are discrepancies. 

Regarding Dataset A, Tables \ref{tab:allStatisticsLandmarks} and \ref{tab:RegionsRanking} shows how we obtain the best ranking results with the region-based approach, but according to the $C_{llr}$ the best results are obtained when using the landmark-based approach with the landmark set 1. This occurs because the $C_{llr}$ also takes into account misleading LR values, penalizing comparisons that, although obtaining the top position in the ranking have excessively high scores.

This also happens, but in favor of the region-based approach, when looking at the experiments with dataset B, where the latter approach, obtains an average ranking of $1.24$ and a maximum ranking of $7$, worse results compared to the landmark-based approach results, but obtains a better $C_{llr}$.

Regarding the complete dataset, if we analyze the ranking results by comparing Tables \ref{tab:allStatisticsLandmarks} and \ref{tab:RegionsRanking}, and using Figure \ref{cmcCurvesALL}, in the case of the region-based approach we obtain an average ranking of $1.5$, slightly better than the $1.6$ obtained by the landmark-based approach when we use the landmark set 1. However, comparing the maximum ranking of the same two experiments, we identify that in the worst case for the region-based approach there is a need to look at $26$ positions in the ranking to find the correct comparison, while for the landmark-based approach only $9$ comparisons are required. In contrast, the region-based approach is able to have more than $95\%$ of cases with a ranking position of $1$, while the landmark-based approach have less than $75\%$ of cases with a ranking position of $1$.

Both approaches achieve very promising results and speed up the process of searching candidates. While the landmark-based approach offers a more accurate method for the most difficult cases, the region-based approach is better at solving many similar cases but worse at solving the most difficult ones. In this comparison of methods, $C_{llr}$ also tells us how much information we need to determine the true hypothesis without prior information. As expected, in the experiment with all the cases, the best approach according to this metric is the region-based approach, since it uses a more informed metric.

\section{Discussion and conclusions} \label{discussion}
\noindent

Forensic odontology plays a key role in the human identification process, being considered a primary identification method by Interpol. Despite recent technological advances, methods for performing morphological comparisons of dentition remain manual and subjective. In this study we have introduced two different approaches to simplify and speed up this task using 3D IOS, an increasingly common type of data in PM scenarios, and photographs where teeth are visible, which are more common and easier to obtain, especially in complex scenarios or when medical records are not accessible or nor-existent. These methods, using computer vision and optimization techniques, have proven to be very useful for ranking candidates.

With the first approach, using paired landmarks to superimpose 3D models and photographs, excellent results have been achieved. Three different sets of landmarks have been tested to find a set that is both simple and useful for the superimposition task. The landmark set 1, with landmarks on the smile line, midline, and gingival line, was able to rank the candidates so that in the worst case, in a varied dataset of $142$ subjects ($20,164$ comparisons in total), the correct comparison can be found in the first $9$ positions, and in the average case only $1.6$ positions are needed. Using the $C_{llr}$, we also see that this system is capable of making the correct decisions with a low expected error rate, with a value of $0.29$ for the experiment mentioned, much lower than $1$, a reference system that does not provide any information.
 
With the second approach, using teeth segmentation with a mask for areas with possible occlusion or gingival problems, excellent results have also been achieved. This approach is capable of sorting cases in such a way that, on average, only $1.5$ positions in the ranking are needed to find the correct match in the dataset with the highest number of cases. Although it is true that for the most difficult case the ranking position is $26$, it performs better in all the other cases. By using a segmentation of the teeth and thus using more information than the landmark-based approach, this approach achieves a better $C_{llr}$, with a value of $0.2316$, showing how reliable are the results. These $C_{llr}$ values (from $0.122$, when the entire teeth are visible, to $0.231$, with limited visibility) are competitive to established identification systems \cite{van_lierop_overview_2024}: a) an Automated Fingerprint Identification System, $C_{llr} = 0.165$; b) one of the best performing voice comparison system, $C_{llr} = 0.207$; or c) one of the best performing facial recognition systems using photographs, $C_{llr} = 0.104$.

The results of this study show very promising applications of computer vision in the field of forensic odontology, but it does have certain limitations. The sample size consists of $142$ cases from a single population. A larger and more varied sample could help validate the results and obtain more reliable LR. Periodontal problems can affect the proposed method, and need to be studied in more detail. The time between AM and PM data collection is between one and three years, and no significant differences in dentition due to dental treatments or problems were observed. This means that we do not know the reliability of the method in cases where the dentition undergoes morphological changes. 

As future work we will study the addition of dental records information, describing the condition of each tooth, so we can identify morphological changes, allowing us to leave the affected areas out of the comparison and thus avoiding possible errors. We also plan to develop methods to automatize the data processing in order to obtain both landmarks and segmentations needed to apply the proposed methods. Also as a future work, we plan to compare 3D intraoral scans with panoramic X-rays scan of the teeth. Panoramic dental X-rays show the entire set of teeth, allowing us to compare the whole dentition. This entails some difficulties, as the system for acquiring a panoramic dental image is much more complex, but it would be a major step forward in terms of being able to make a more complete morphological comparison.

\section*{Acknowledgments}
\noindent

This publication is part of the R\&D\&I project PID2024-156434NB-I00 (CONFIA2), funded by MICIU/AEI/10.13039/501100011033 and ERDF/EU. This work is also funded by ‘EIC Accelerator - Seal of Excellence’ project (09/942572.9/23) within the 2023 call for aid from the Community of Madrid to finance projects that have obtained a Seal of Excellence within the European Innovation Council’s Accelerator Program, and by CESPU—Cooperativa de Ensino Superior Polit\'ecnico e Universit\'ario under the grant MLIA\_REAB-GI2-CESPU-2025.

Dr. Ib\'a\~nez's work is funded by the Spanish Ministry of Science, Innovation and Universities under grant RYC2020-029454-I and by Xunta de Galicia by grant ED431F 2022/21. We wish to acknowledge the support received from the Centro de Investigación de Galicia ``CITIC'', funded by Xunta de Galicia and the European Union (ERDF- Galicia 2014-2020 Program), by grant ED431G 2019/01.

Xavier Abreu-Freire benefited from an Erasmus+ mobility grant for SMP – Traineeships (ref. 2024-1-PT01-KA131-HED-000196087-012) during the period in which he collaborated on this study at Panacea Cooperative Research.

Xavier Abreu-Freire and Daniel P\'erez-Mongiovi were supported by FCT - Funda\c c\~ao para a Ci\^encia e Tecnologia, I.P., in the scope of the project UID/04378/2025 (10.54499/UID/04378/2025), and UID/PRR/04378/2025 (10.54499/UID/PRR/04378/2025), of the Research Unit on Applied Molecular Biosciences - UCIBIO and the project LA/P/0140/2020 (10.54499/LA/P/0140/2020) of the Associate Laboratory Institute for Health and Bioeconomy - i4HB.

This project was made possible through the access granted by the Galician Supercomputing Center (CESGA) to its supercomputing infrastructure. The FinisTerrae III have been funded by the NextGeneration EU 2021 Recovery, Transformation and Resilience Plan, ICT2021-006904, and also from the Pluriregional Operational Programme of Spain 2014-2020 of the ERDF, ICTS-2019-02-CESGA-3, and from the State Programme for the Promotion of Scientific and Technical Research of Excellence of the State Plan for Scientific and Technical Research and Innovation 2013-2016, CESG15-DE-3114.

\printbibliography

\newpage

\vfill

\end{document}